\documentclass[letterpaper, 10pt, conference]{ieeeconf}
\IEEEoverridecommandlockouts                              
\usepackage{graphics, placeins}
\usepackage{graphicx}
\usepackage{subcaption}
\usepackage{epsfig}
\usepackage{mathptmx}
\usepackage{times}
\usepackage{lipsum}
\usepackage{subcaption}
\usepackage[hidelinks]{hyperref}
\usepackage{amssymb, amsmath, amsfonts, newtxmath}
\usepackage{cancel}
\usepackage{algorithm}
\usepackage{algpseudocode}
\usepackage{algorithmicx}
\usepackage{dsfont}

\algblock{ParFor}{EndParFor}
\algnewcommand\algorithmicparfor{\textbf{parfor}}
\algnewcommand\algorithmicpardo{\textbf{do}}
\algnewcommand\algorithmicendparfor{\textbf{end\ parfor}}
\algrenewtext{ParFor}[1]{\algorithmicparfor\ #1\ \algorithmicpardo}
\algrenewtext{EndParFor}{\algorithmicendparfor}

\usepackage[backend=biber,bibencoding=utf8, style=numeric-comp , autocite=plain, sorting=none]{biblatex}
\graphicspath{{./images/} {./}}

\newcommand{\blockcomment}[1]{\iffalse #1 \fi}

\title{\LARGE \bf Enhancing Robotic System Robustness via\\Lyapunov Exponent-Based Optimization}
\author{
    G. Fadini$^{1,\star}$,
    S. Coros$^{1}$,
    \thanks{$^{1}$ETHZ, Computational Robotics Lab (CRL), Z\"urich, Switzerland}
    \thanks{$^\star$ Corresponding author: \href{mailto:gfadini@ethz.ch}{\tt{gfadini@ethz.ch}}}
}

\begin{document}

        \maketitle
        \thispagestyle{empty}
        \pagestyle{empty}

        \begin{abstract}
            We present a novel approach to quantifying and optimizing stability in robotic systems based on the Lyapunov exponents
            addressing an open challenge in the field of robot analysis, design, and optimization.
            Our method leverages differentiable simulation over extended time horizons.
            The proposed metric offers several properties, including 
            a natural extension to limit cycles commonly encountered in robotics tasks and locomotion.
            We showcase, with an \textit{ad-hoc} JAX gradient-based optimization framework,
            remarkable power, and flexibility in tackling the robustness challenge.
            The effectiveness of our approach is tested through diverse scenarios of varying complexity,
            encompassing high-degree-of-freedom systems and contact-rich environments. The positive outcomes across these cases highlight the potential of our method in enhancing system robustness.
        \end{abstract}

    \section{Introduction}
    Morphological computation \cite{muller_what_2017}, a biologically-inspired concept, could 
    enhance robotic design by offloading computational tasks to the physical body,
    thereby reducing reliance on onboard processing and control corrections.
    If combined with co-design strategies, this approach allows for the simultaneous optimization of a robot's physical structure and control logic,
    resulting in systems that are robust and efficient in complex, dynamic environments.
    For instance, a legged robot could navigate uneven terrain using the same control signals as on flat ground, relying on its mechanical structure to gracefully adapt to perturbations.
    Embodied intelligence would lead to more natural and robust behaviors, 
    promising to advance robotic capabilities in unpredictable scenarios.
    However, it is difficult to find a metric that describes this kind of resilience to perturbation.
    While it is clear how to score the performance of a robot while performing a task or
    to draw conclusions about its hardware design, it is rather unclear what tools to use
    to concretely assess its ultimate robustness. 
    The design of robust robotic systems for unplanned scenarios remains still an open problem.
    To address such a question, in this paper, we propose to tackle the problem
    of finding a robustness metric, with the ultimate goal of robot co-design in mind.
    Prior research in this field has primarily focused on optimality \cite{fadini_computational_2021, fadini_unpublished, ha_computational_2018, geilinger_skaterbots_2018}.
    However, optimal trajectories may not always be feasible in real systems due to unmodelled dynamics,
    noise, delays, saturation, or actuator dynamics that can hinder effective perturbation rejection.
    While optimality remains crucial, robustness emerges as a complementary aspect necessary for real-world deployment.
    Current work aims to select trajectories and designs requiring minimal control correction on real hardware.
    Proposed strategies include stochastic optimization \cite{bravo-palacios_robust_2022, palacios2022} and data-driven approaches \cite{fadini2022},
    where a robust bi-level scheme incorporating additional simulations to enhance co-design robustness was proposed, tailoring hardware parameters
    for robustness.
    However, these approaches faced scalability issues and relied on proxies depending on the
    environment and noise injection.
    Other research \cite{maywald_co-optimization_2022, ghirlanda2024} explored maximizing the region of attraction for
    stabilizing controllers in simple underactuated systems.
    For parametric optimization, \cite{giordano_trajectory_2018} outlined a method for selecting optimal trajectories for UAVs using sensitivity analysis. However, the applicability of this method to legged robots with more degrees of freedom and switching contact dynamics requires careful reconsideration.
    To extend similar results to legged robotics,
    gradients could be obtained via differentiable simulation \cite{geilinger_add_2020, suh_differentiable_2022},
    although challenges arise due to the non-smooth nature of the problem.
    
    As a foundation to future work in this direction,
    we propose a method that assesses the sensitivity of a robotic system to perturbations
    and that can be hence used in several analysis or optimization scenarios.
    Contact-rich loco-manipulation problems and some dynamical systems show the property of deterministic chaos.
    Our formulation's advantage is offering a clearer theoretical link with
    the properties of their non-linear dynamics.
    The Lyapunov exponents, which are at the cornerstone of our method,
    give a powerful tool to characterize non-linear systems long-term behavior
    \cite{Kunze2000, Barreira2017}.
    Their use is well established for dynamical systems and
    previous literature speculated about their possible use it in robot hardware 
    optimization to increase stability \cite{zang_applications_2016}.
    This concept has also been already used to evaluate robustness in simplified bipedal systems
    \cite{yang_stabilization_2006, yunping_stability_2013},
    offering potential insights into robustness metrics in legged robotics.
    However, it has seen little use in other systems with high dimensionality and in gradient-based optimization.
    To fill this gap, we propose to include a differentiable formulation adapted to
    several applications, including system analysis, optimization, and differentiable co-optimization of system and plant.

    \section{Mathematical background}
    \label{sec:background}
    \subsection{Sensitivity analysis in discrete dynamical systems}
    In our analysis, systems with discrete dynamics are considered.
    Equivalent considerations can also be extended to the case of systems with continuous dynamics.
    In the case of Markovian autonomous systems, it is possible to discretize the dynamics via a transition map
    $\Phi$ that links each state to the next one. This dependency is written as:
    \begin{equation}
        x_{i+1} = \Phi(x_i)
    \end{equation}
    Intuitively $\Phi$ can be interpreted as a single forward simulation step which links one
    state $x_i$ to the next one $x_{i+1}$.
    In a complete simulation rollout,
    the state transition map is applied iteratively for $n$ steps starting from the initial condition $x_0$.
    This leads to the following expression of the final state $x_n$:
    \begin{equation}
      x_{n} = \underbrace{\Phi(\Phi( \dots \Phi}_{\text{{n times}}}(x_0)))
      \label{eq:iteration}
    \end{equation}
    The final state sensitivity can be computed with respect to the initial state
    with a telescopic expansion as:

    \begin{equation}
      \dfrac{\partial x_n}{\partial x_0} = 
      \underbrace{
      \left.\dfrac{\partial \Phi}{\partial x}\right|_{x_{n-1}}
      }_{\text{d} \Phi(x_{n-1})}
      \left.\dfrac{\partial \Phi}{\partial x}\right|_{x_{n-2}}
      \dots
      \left.\dfrac{\partial \Phi}{\partial x}\right|_{x_{1}}
      \left.\dfrac{\partial \Phi}{\partial x}\right|_{x_{0}}
      =
      \displaystyle{\Pi}_{i=0}^{n-1} \text{d} \Phi(x_i)
      \label{eq:iterative_form}
    \end{equation}
    The sensitivity magnitude of the final state to the initial condition
    can be found after applying the
    transition map $\Phi$ for $n$ times can be obtained by \eqref{eq:iterative_form}.

    \subsection{Divergence of infinitesimally perturbed trajectories}
    Two trajectories $\Gamma$ and $\Gamma'$ starting respectively from $x_0$ and $x_0'$ are considered,
    $x_0'$ differs from $x_0$ of an arbitrary small quantity $\varepsilon$ at the initial state $x_0$,
    that is $x'_0 = x_0 + \varepsilon$.
    These trajectories will evolve under the transition map $\Phi$ as:
    \begin{equation}
        \begin{split}
            \Gamma = \{x_0, x_1 = \Phi(x_0), \dots x_i \dots \}\\
            \Gamma' = \{ x'_0, x'_1 = \Phi(x'_0), \dots x'_i \dots \}\\
        \end{split}
    \end{equation}
    For linearity of $\text{d} \Phi$, under small perturbations $\varepsilon \approx 0$,
    the steps can be compounded. Neglecting higher-order terms, the propagation of the perturbation
    at the i-th iteration is:
    \begin{equation}
        x'_i - x_i = \Pi_{k=0}^{i-1} \text{d}\Phi(x_k) \varepsilon = \Pi_{k=0}^{i-1} \text{d}\Phi(x_k) (x'_0 - x_0)
    \end{equation}
    \begin{figure}
        \centering
        \includegraphics[width=0.9\linewidth]{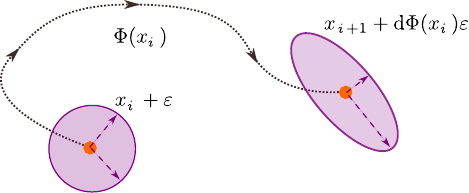}
        \caption{The transition map $\Phi$ maps an uncertainty of radius $\varepsilon$ around $x_i$
        into an ellipsoid $\varepsilon^\top \text{d}\Phi^\top(x_i) \text{d}\Phi(x_i) \varepsilon$.
        }
        \label{fig:covariance_propagation}
    \end{figure}

    So $\text{d} \Phi(x_n)$ is tied to the variations that
    a ball ${\mathcal{B}_\varepsilon(x_i) \overset{\Delta}{=} \{x\in \mathbb{R}^{n} \text{ such that } \|x - x_i\| < |\varepsilon|\}}$
    of phase space around point $x_n$ undergoes
    after applying the transformation $\Phi$, this is visualized in Fig.~\ref{fig:covariance_propagation}.
    The eigenvalues of $\text{d}\Phi(x_i)$ correspond to the magnitude of the
    deformation along the principal axis of the linear transformation around $x_i$.
    Their product gives an information about the transformation:
    \begin{itemize}
        \item For $|\text{d} \Phi(x_i)| > 1$ the volume is expanding.
        \item For $|\text{d} \Phi(x_i)| < 1$ the volume is contracting.
        \item For $|\text{d} \Phi(x_i)| = 1$ the volume is conserved.
    \end{itemize}
    
    \subsection{Lyapunov exponent definition}
    For linear systems, $\text{d} \Phi$ is constant at any iteration, so studying it once is enough
    to draw conclusions about the convergence of the system's trajectories. 
    Conversely, in the case of non-linear systems, $\text{d}\Phi$ depends on the state,
    and the evolution of the whole trajectory is necessary to have an assessment.
    The analysis of the convergence property is recovered by averaging the
    transformations along the whole trajectory.
    The Lyapunov exponents are a tool to characterize the long-term behavior of nearby trajectories in non-linear systems \cite{arnold1986lyapunov,
        eckmann1985ergodic,
        ginelli2007characterizing,
        grassberger1983characterization,
        wilkinson2016what,
        datseris_nonlinear_2022,
        }.
    For one-dimensional maps, the Lyapunov exponents are computed as the time average of $\log |\text{d}\Phi|$.
    This concept can be extended to higher-dimensional maps and flows
    thanks to Oseledet's ergodic multiplicative theorem \cite{Oseledets1968}.
    They are defined as the limit of the logarithm of the geometric average of the transformations for
    infinitely long trajectories:
    \begin{equation}
        \lambda = \lim_{n \rightarrow \infty} \frac{1}{n} \ln{\| \Pi_{i = 0}^{n} \text{d} \Phi(x_{i})\|}
        \label{eq:lyapunov_exponent}
    \end{equation}
    In these cases, there are multiple exponents, equal in number to the dimension of the phase space.
    These exponents are typically arranged in descending order.
    The generalization to higher dimensions provides a powerful tool for analyzing the asymptotic behavior of
    complex dynamical systems, offering insights into how nearby trajectories diverge or converge over time
    in different directions of the phase space.
    \subsection{Computation of the Lyapunov exponents}
    Numerical techniques can approximate \eqref{eq:lyapunov_exponent} for multidimensional systems and some libraries
    are already readily available \cite{Datseris2018}.
    A common algorithm uses QR decomposition \cite{datseris_nonlinear_2022} so that at any instant $i$:
    $\text{d}\Phi_i = Q_i R_i$ where $Q_i$ is an orthonormal matrix and $R_i$ upper triangular matrix.
    The $j$-th Lyapunov exponent is then computed by:
    \begin{equation}
        \lambda_j = \frac{1}{N \Delta t} \sum_{i=0}^{N} \log(|R_{i[j,j]}|)
        \label{eq:qr_lyap}
    \end{equation}
    where $\Delta t$ is the time discretization between two states. 
    As in floating base robotics, it may happen that $\text{d}\Phi_i$ is not a square matrix applying \eqref{eq:qr_lyap} is not viable.
    Our approach to solving this is to consider the matrix $\text{d}\Phi_i^{\top} \text{d}\Phi$ instead.
    Moreover, we employ a singular values decomposition so that ${\text{d}\Phi_i^{\top} \text{d}\Phi_i = U_i \Sigma_i V_i}$.
    Where $U_i$ is the matrix of the left singular vectors, $V_i$ is the matrix of the right singular vectors and
    $\Sigma_i$ is a diagonal matrix containing the squared eigenvalues on its diagonal. 
    SVD is differentiable \cite{Zhang2019AutomaticDC}, and despite having a higher computational cost compared to QR decomposition,
    it is better suited for near-singular cases.
    \begin{equation}
        \lambda = \frac{1}{2 N \Delta t} \sum_{i=0}^{N} \log(\Sigma_{i})
        \label{eq:qr_lyap}
    \end{equation}

    \subsection{Properties of the Lyapunov exponent}
    The Lyapunov exponents do not depend on the initial conditions
    and are invariant with respect to coordinate changes (as only
    the information about the magnitude of the singular values of the transformation $\Phi$
    is used).
    The Lyapunov exponent is a characteristic of the system and its asymptotic behavior.
    For any initial point in the state-space, whose evolution tends to a given
    invariant set, the same value of the Lyapunov exponent is obtained, as we show next.

    \paragraph*{Numerical example: Van der Pol oscillator}
    \begin{figure}
        \centering
        \includegraphics[width=\linewidth]{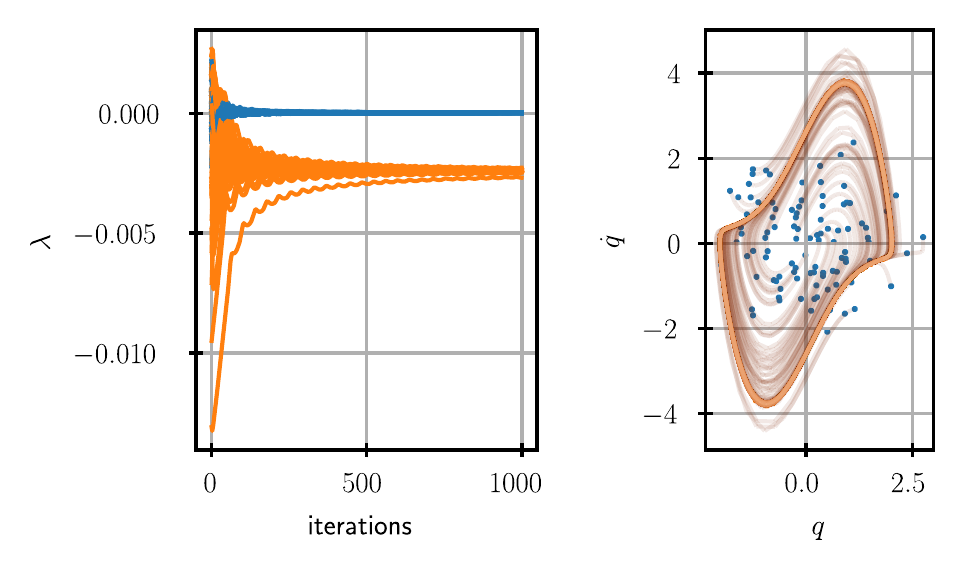}
        \caption{Van der Pol oscillator $\mu=2$, on the left the Lyapunov spectrum is approximated for
        100 different starting conditions.
        On the right, the trajectories in the
        state space are shown (starting points in blue). The shape of the attractor can be seen in light orange.}
        \label{fig:van_der_pol}
    \end{figure}
    A non-linear oscillator \cite{vanderPol1920} is taken to exemplify these properties.
    Its continuous dynamics is governed by
    $
    {
        \ddot{q} - \mu (1-q^2)\dot{q} + q = 0
    }
    $. We discretize it with an Euler integrator and a timestep $\Delta t=10^{-3}s$ and we compute the
    Lyapunov exponents approximation 
    with an increasing number of iterations under a total simulation time of $100s$.
    The computation of the trajectories and $\lambda$ is performed for
    100 different initial conditions. 
    In Fig.~\ref{fig:van_der_pol} the results are plotted, on the left we show the approximation
    of the two components of $\lambda$ by increasing the iterations (simulation steps),
    while on the right the evolution of the trajectories is shown in the phase plane. 
    %The limit $n \rightarrow \infty$ makes the information of the initial conditions less and less important.
    The asymptotic convergence of $\lambda \approx [0, -2\cdot 10^{-3}]$ shows that the memory of the initial state
    is progressively lost as the number of simulation steps increases.
    The presence of a zero in $\lambda$ means that along one dimension of the attractor, the trajectories are neither
    diverging nor converging, while the negative value can be interpreted by the fact that in the other direction,
    the trajectories are being attracted to the limit cycle. 
    This example shows that for ergodic systems, the values of $\lambda$ are invariant properties provided that
    the trajectories converge to the same attractor
    (either being an equilibrium point or a limit cycle,
    as in this case) and that the simulation is long enough.
    This is a very useful property when the long-term stability of quasi-periodic motions is to be studied.

    \section{Proposed computational framework}
    \label{sec:framework}
\begin{figure}
        \centering
        \includegraphics[width=0.75\linewidth]{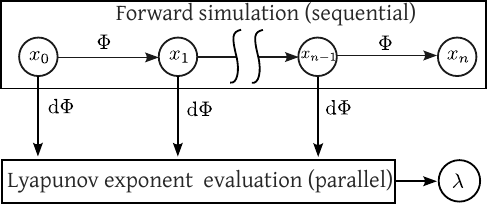}
        \caption{Parallelized Lyapunov spectrum computation: first a forward simulation is
        performed and then the computation of the spectrum is parallelized with respect to the states
        using the simulator gradients.}
        \label{fig:lyapunov_compute}
    \end{figure}

    The algorithm to compute $\lambda$ is shown in Fig.~\ref{fig:lyapunov_compute}
    and explained in Algorithm 1.
    A forward simulation allows to produce a trajectory rollout $\Gamma$, then the
    computation and manipulation of $\text{d}\Phi$ is parallelized for each state.
    \begin{algorithm}
        \caption{Lyapunov exponent computation}
        \begin{algorithmic}[1]
        \Require Initial state $x_0$, simulation steps $N$, timestep $\Delta t$
        \State Initialize $x \gets x_0$
        \For{$i = 1$ to $N$}
            \State $x_{i} \gets \Phi(x_{i-1})$ \Comment{Forward simulation step}
        \EndFor
        \ParFor{$i = 1$ to $N$} \Comment{Parallel computation}
            \State $U_i,\Sigma_i,V_i \gets \text{SVD}(\text{d}\Phi(x_i)^{\top}\text{d}\Phi(x_i))$ \Comment{Get spectrum}
            \State $\lambda_i \gets \text{log}(\text{diag}(\Sigma_i))$ \Comment{Log transformation} 
        \EndParFor
        \State $\lambda \gets \text{sum}(\lambda_{0}, \lambda_1, \ldots, \lambda_{N-1})/(2 N \Delta t)$ \Comment{Combine results}
        \State \Return $\lambda$
        \end{algorithmic}
    \end{algorithm}
    After the intermediate operations are carried out,
    the final estimate of $\lambda$ is achieved and the stability of the system can be assessed.
    For the selection of our robustness metric,
    we remark that the sum of the components of $\lambda$ is connected to the deformations
    that a hypervolume in the state space undergoes with time.
    For robustness, we identify the system property to possess trajectories which are not
    diverging the one from the others, so, given these results,
    as a metric of robustness $\mathcal{L}_{\lambda}$,
    the signed sum of the Lyapunov exponent components is taken:
    \begin{equation}
        \mathcal{L}_{\lambda} = \mathds{1}^{\top} \lambda
    \end{equation}
    where : $\mathds{1}$ is the ones vector of the same dimension of $\lambda$.
    In Hamiltonian systems, where energy is conserved, no overall deformation of
    a state space volume is observed according to Liouville's theorem \cite{Phillips1969}, and then
    $\mathcal{L}_{\lambda}=0$.
    Conversely, in dissipative systems, when the system is asymptotically converging to a stable equilibrium point, then
    volumes of state space are shrinking under $\Phi$, leading to $\mathcal{L}_\lambda < 0$.
    Finally, in chaotic systems, positive Lypaunov exponents appear, meaning
    that system trajectories are diverging along one direction of the linearization along the trajectory. 

    \subsection{Extension to robotic systems}
    \begin{figure}[ht]
        \centering
        \includegraphics[width=0.9\linewidth]{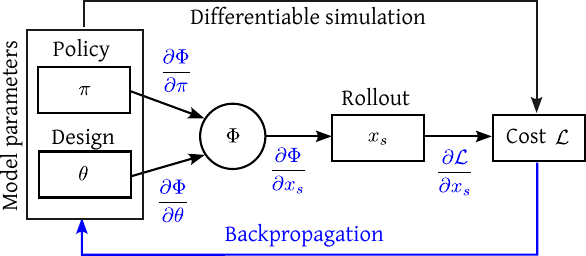}
        \caption{Framework for co-optimization of hardware and policy parameters with
        differentiable simulation.}
        \label{fig:framework}
    \end{figure}
    In the case of actuated robotic systems the state transition function also includes some form of control action:
    \begin{equation}
        x_{i+1} = \Phi(x_i, u_i)
        \label{eq:state_transition}
    \end{equation}
    Where $x$ represents the state, $u$ is the control and $i$ is the index associated with a trajectory state. 
    In this work, we assume the existence of a controller $\Pi$, so that ${u_i = \Pi(x_i)}$.
    Such control policy $\Pi$ can be integrated into the transition map \eqref{eq:state_transition} 
    making the system autonomous and thus the theory explained so far is applicable.
    In Fig.~\ref{fig:framework} the workflow of our differentiable physics approach is shown.
    The robot's dynamics $\Phi$ is formulated with a differentiable physics simulator,
    possibly including contact interactions, allowing for the computation of gradients with respect
    to both hardware and control parameters.
    A differentiable cost function $\mathcal{L}$ is formulated to capture the desired performance and robustness, including $\mathcal{L}_\lambda$.
    During a forward pass, the simulator runs with the current parameters,
    generating a trajectory of states and evaluating such costs.
    Backpropagation can also be employed to compute the gradients of the loss function from the current state.
    These gradients are propagate back through the simulation to possibly
    update hardware or control parameters using first order optimization algorithms.
    Forward simulation and backpropagation updates can then be repeated to increase robustness.
    This pipeline is adapt for a holistic optimization of robotic systems via
    controller and plant co-design, ultimately leading to more efficient and capable robotic design and control.
    In practice, however, we can also just selectively focus on specific components of the system.
    Via a forward rollout, the robustness of the current model can be quantified.
    Via a hardware model update, the design can be improved so it better performs under a given controller.
    Finally, via a policy update, the control policy can be refined to increase robustness.
    
    \section{Results}
    \subsection{Implementation}
    In this section, we apply the metric defined in the previous chapter to 
    several robotics case studies.
    % two co-design optimizations, one
    % with a robotic manipulator and the second with a quadruped robot.
    % For the synthesis of a fall-recovery policy and a simple locomotion one.
    To write the dynamics the MJX module of \cite{todorov2012mujoco} was used.
    This simulator enables to computation of trajectory rollouts and the derivative $\text{d}\Phi$.
    Thanks to it we could also obtain intermediate JAX \cite{jax2018github} representation which could be differentiated
    and parallelized as explained in the previous section.

    \subsection{Planar manipulator - controller \& plant co-design}
    \begin{figure}
        \centering
        \includegraphics[width=0.65\linewidth]{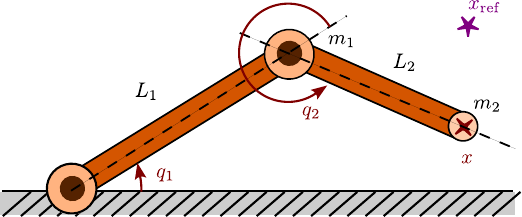}
        \caption{Robot manipulator scheme}
        \label{fig:manipulator}
    \end{figure}
\begin{figure}[htbp]
    \centering
    \begin{subfigure}{0.45\linewidth}
        \centering
        \includegraphics[width=\linewidth]{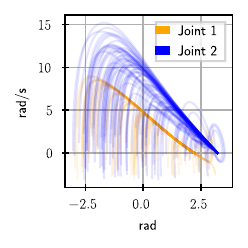}
        \caption{Optimized state-space}
        \label{fig:mani_ss_opti}
    \end{subfigure}
    \begin{subfigure}{0.45\linewidth}
        \centering
        \includegraphics[width=\textwidth]{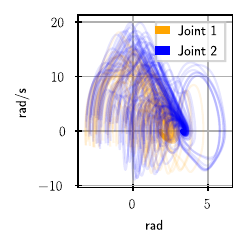}
        \caption{Nominal state-space}
        \label{fig:mani_ss_nomi}
    \end{subfigure}
    \begin{subfigure}{0.45\linewidth}
        \centering
        \includegraphics[width=\textwidth]{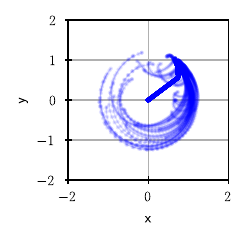}
        \caption{Optimized cartesian}
        \label{fig:mani_car_opti}
    \end{subfigure}
    \begin{subfigure}{0.45\linewidth}
        \centering
        \includegraphics[width=\textwidth]{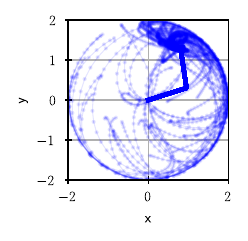}
        \caption{Nominal cartesian}
        \label{fig:mani_car_nomi}
    \end{subfigure}
    \caption{The two plot on the left show 100 trajectories starting from
        random initial states in both the state space and
        the cartesian space for optimized hardware and policy,
        while the ones on the right show 100 random trajectories with nominal parameters.}
        \label{fig:manipulator_res}
    \end{figure}
\begin{figure}
        \centering
        \includegraphics[width=0.8\linewidth]{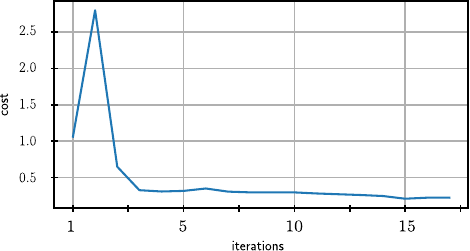}
        \caption{Cost trend during the co-optimization of the manipulator. The cost reaches a plateau after around 5 iterations.}
        \label{fig:manipulator_cost}
    \end{figure}

    We consider a planar manipulator as shown in Fig.~\ref{fig:manipulator} made up by:
    two actuated revolute joints controlling the DoFs $q_1$ and $q_2$ and two rigid links with lengths $L_1, L_2$
    with concentrated masses $m_1$ and $m_2$.
    The actuator is influenced by a downward gravitational force.
    As hardware parameters to optimize we select:
        the link lengths $(L_1, L_2)$ and the two concentrated masses $(m_1, m_2)$.
    The task we select is to stabilize the manipulator reaching a target position ${x_{\text{ref}}=[0.7, 0.7]}$ with the end effector $x$.
    In order to do so we select as control policy $\Pi$ a PD controller with cartesian stiffness \cite{hogan_impedance_1985}:
    ${
            \Pi(x_i) = \text{K}_p (q_i-q_\text{ref}) + \text{K}_d (v_i) + 
                    \text{K}_c J^\top(q_i) (x -x_\text{ref})%+\\
      }
    $.
    The system is simulated for 2s with a $\Delta t = 10^{-3}$.
    As co-optimization variables, we select the lengths of the links, the moving masses
    that impact the inertia of the system, and the policy gains.
    The optimization relies on the ADAM optimizer \cite{Kingma2014AdamAM} from the Optax library \cite{deepmind2020jax}.
    The cost convergence plot is shown in Fig.~\ref{fig:manipulator_cost} where a plateau is reached after around 5 iterations.
    The results are shown in Tab.~\ref{tab:comparison}, the metric of the robustness of the optimized robot $\mathcal{L}_{\lambda}$ is
    over 5 times higher than the nominal one. 
    In Fig.~\ref{fig:manipulator_res} the nominal manipulator (right plots) and optimized one (left plots) are tested by starting the simulations from 100 random joint positions and velocities.
    The optimized system shows trajectories (Fig.~\ref{fig:mani_car_opti}) that converge more steadily to the final position in cartesian space than
    the nominal (Fig,~\ref{fig:mani_car_nomi}).
    We also observe that selecting the optimized case
    produces trajectories that converge faster to the equilibrium point and are more regular in the state space (Fig.~\ref{fig:mani_ss_opti}) if compared to nominal (Fig.~\ref{fig:mani_ss_nomi}).
    \begin{table}[tbp]
        \caption{Comparison of nominal and optimized parameters}
        \label{tab:comparison}
        \centering
        \begin{tabular}{|l|c|c|}
            \hline
            \textbf{Parameter} & \textbf{Nominal} & \textbf{Optimized} \\
            \hline
            Lengths $L_1, L_2$ [m] & (1.0, 1.0) & (0.9, 0.1) \\
            \hline
            Masses $m_1, m_2$ [kg] & (0.5, 0.5) & (0.9, 0.01) \\
            \hline
            Damping $K_d$ [Ns/rad] & (5.0, 5.0) & (31.7, 21.7) \\
            \hline
            Stiffness $K_p$ [Nm/rad] & (20.0, 20.0) & (59.8, 59.8) \\
            \hline
            Stiffness $K_c$ [Nm/rad] & (1.0, 1.0) & (0.75, 1.78) \\
            \hline
            \hline
            \textbf{Cost} & 1.06 & 0.21 \\
            \hline
            \textbf{Robustness $\mathcal{L}_{\lambda}$} & $-2.5\cdot 10^{-3}$ & $-1.3\cdot 10^{-2}$\\
            \hline
        \end{tabular}
    \end{table}
    
    \subsection{Falling spider robot - assessing system robustness}
\begin{figure}[tbp]
    \centering
    \begin{subfigure}[c]{0.4\linewidth}
        \centering
        \raisebox{-0.5\height}{
        \includegraphics[width=\linewidth]{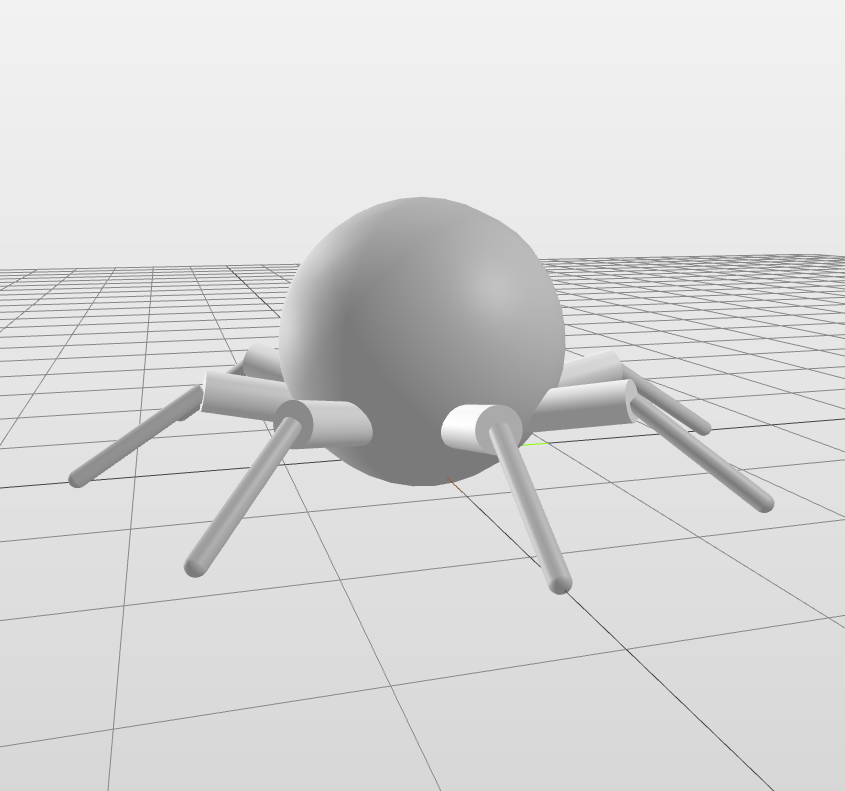}
        }
        \hfill
        \caption{Spider-like robot}
        \label{fig:spider_mod}
    \end{subfigure}
    \begin{subfigure}[c]{0.55\linewidth}
        \centering
        \includegraphics[width=\linewidth]{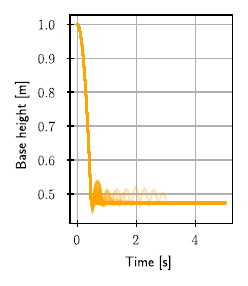}
        \caption{Simulations}
        \label{fig:spider_sim}
    \end{subfigure}

    \caption{Model of the 16 DoF spider \ref{fig:spider_mod} and 100 simulation rollouts \ref{fig:spider_sim} with different combinations
        of the joint controller PD gains and same position reference.
        The final height reaches steady values after a quick transient, as energy
        is dissipated in the joints and with ground interaction.}
        \label{fig:spider}
    \end{figure}

\begin{figure}
        \centering
        \includegraphics[width=0.95\linewidth]{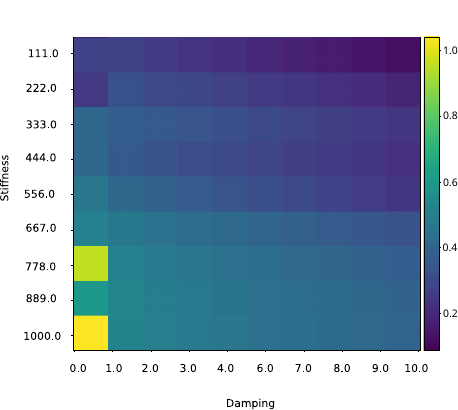}
        \caption{Trend of the robustness metric $\mathcal{L}_{\lambda}$ for different combinations of the joint controller gains parameters (normalized)}
        \label{fig:spider_lyap}
    \end{figure}

    A poly-articulated robot's fall dynamic is studied to gain insights
    into the relationship between controller parameters and the metric of robustness
    we propose.
    We test the computation of $\mathcal{L}_{\lambda}$ for different combinations of a joint
    PD controller and fixed design parameters. As a test case the spider-like robot made up of 16
    joints and a floating base as in Fig.~\ref{fig:spider}.
    We make the robot fall from the same initial condition at a height of $1m$ as can be seen in
    Fig.~\ref{fig:spider_sim}.
    By computing $\mathcal{L}_{\lambda}$ we aim to evaluate the robot's behavior under varying joint controller gains
    (stiffness and damping). These results are reported in Fig.~\ref{fig:spider_lyap}.
    There we can observe that lower robustness values were consistently associated with more heavily damped systems that exhibited lower joint stiffness.
    This finding suggests that increased damping and reduced stiffness contribute to increased robustness during the robot's fall.
    Such a setting likely allows the system to absorb and dissipate energy faster, hence resulting in a more controlled motion.
    This outcome aligns with intuitive expectations, as a more dissipative system would generally be better equipped
    to handle the impact and unpredictability associated with falling.
    Such a result further consolidates the use of the metric as a valuable analysis tool
    for tuning control parameters in robotic systems to enhance their resilience to unexpected disturbances.

    \subsection{Crawling quadruped - policy robustness evaluation}

\begin{figure}
        \centering
        \includegraphics[width=0.5\linewidth]{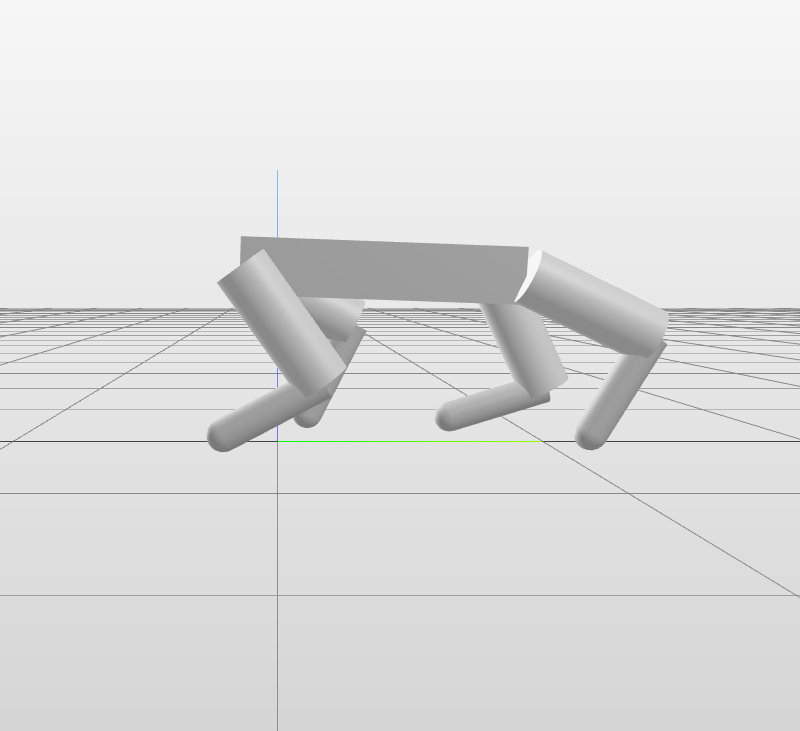}
        \caption{Quadruped robot}
        \label{fig:quadruped-robot}
    \end{figure}

    In this final study, we investigate a locomotion problem with a quadruped robot and
    a PD controller with fixed design parameters.
    The robot model consists of 8 actuated joints and a floating base, as illustrated in Fig. \ref{fig:quadruped-robot}.
    We focus our efforts on evaluating robustness and tuning the controller parameters.
    We aim to achieve a stable and 
    more robust locomotion policy and to assess its robustness.
    Our proxy policy features parametric sinusoidal signals \eqref{eq:signals} superimposed on a constant reference  for each joint.
    To reduce the parameter space, across the joints, we implement shared amplitudes $A_0, A_1$ and frequencies $F_0, F_1$
    and phases $P_0, P_1$ for the different hip and knee joints.
    An additional parameter $\delta$ adds a phase between each pair of opposing legs.
    \begin{equation}
        \begin{aligned}
            q_{\text{\{FR, RL\} HFE}} &= q_\text{ref,hip} + A_0 \sin(2\pi F_0 t + P_0) \\
            q_{\text{\{FR, RL\} KFE}} &= q_\text{ref,knee} + A_1 \sin(2\pi F_1 t + P_1) \\ 
            q_{\text{\{FL, RR\} HFE}} &= q_\text{ref,hip} + A_0 \sin(2\pi F_0 t + P_0 + \delta) \\
            q_{\text{\{FL, RR\} KFE}} &=q_\text{ref,knee} +  A_1 \sin(2\pi F_1 t + P_1 + \delta)
        \end{aligned}
        \label{eq:signals}
    \end{equation}
    \begin{equation}
            \mathbf{u}(t) = k_p (\mathbf{q}_{\text{ref}}(t) - \mathbf{q}) + k_d (\dot{\mathbf{q}}_{\text{ref}}(t) - \dot{\mathbf{q}})
        \label{eq:policy}
    \end{equation}

\begin{figure}
        \centering
        \begin{subfigure}{\linewidth}
            \includegraphics[width=\linewidth]{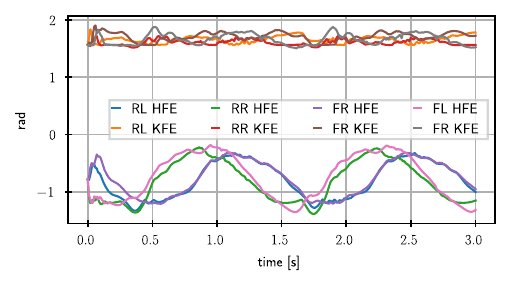}
            \caption{Actuated joints position}
        \end{subfigure}
        \begin{subfigure}{\linewidth}
            \includegraphics[width=\linewidth]{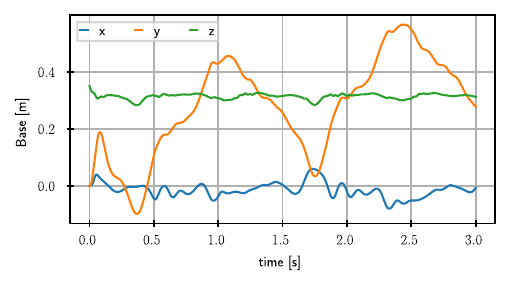}
            \caption{Base position}
        \end{subfigure}
        \caption{Periodic trajectory obtained in simulation with the quadruped robot under the optimized policy.}
        \label{fig:quadruped_trajectories}
    \end{figure}

    The simple policy we obtain \eqref{eq:policy} is parametrized by a total of 9 parameters, including the PD gains to track the position
    and velocity reference. 
    We optimize with JAX and Adam the different parameters in the policy to produce forward motion.
    We simulate for a total time of $10s$ with a $\Delta t = 0.01s$
    In this case, the cost function was left very simple in order to maintain the interpretability of the result.
    Other than the robustness metric, in the loss function
    some conditions on the forward velocity to be close to $0.4m/s$ on the $y$ direction and the height of the base to stay around
    a reference value of $0.3m$ as a regularization.
    The optimized policy results in the crawling trajectories for the joint and the base shown in 
    Fig.~\ref{fig:quadruped_trajectories}.
    Thanks to the differentiable MJX framework, we are able to obtain gradients through the environment
    and this results in a rich contact interaction that exploits different contact modes to produce the motion.
    We compare the the results with the baseline policy and we observe that the value of the optimized policy for robustness
    is $\mathcal{L}_\lambda = -0.61$ while the baseline was $\mathcal{L}_\lambda = -0.56$.
    Both values are negative, indicating the stability of the trajectories around the limit cycle.
    This result constitutes, to the best of the author's knowledge, the first attempt to use the Lyapunov exponents theory for a
    analyzing a non-linear system of this complexity.
    
    \section{Conclusions and future work}
    In this analysis, we explore the quantification of robustness in the context of system optimization, analysis
    and design.
    We propose a differentiable metric,
    which takes into account the non-linear and chaotic behavior of system with contact.
    The key feature of our method is to consider the concept of Lypaunov exponents,
    and minimize its estimated value with a long time horizon leveraging differentiable simulation.
    Using the Lyapunov exponents as a metric, has the advantage to be invariant if a system tends to a specific attractor.
    Hence estimations of this metric can be obtained both for stable equilibria and limit cycles.
    We test this metric to be a good measure of the robustness property for several systems,
    including problems with high degrees of freedom and contact.
    These examples showcase different scenarios, with different aims and complexity.
    The positive outcomes from these diverse cases suggest that our approach holds significant promise
    for enhancing system robustness.
    Moreover, we believe that the potential applications of this methodology extend far beyond this initial investigations.
    For instance, this method could be systematically used to inform and optimize the design process itself,
    potentially leading to inherently more robust systems from the outset.
    With a more based and quantifiable measure of robustness,
    our method opens up new avenues for creating more resilient systems
    in the face of uncertainties and perturbations.
    One of the main challenges in our framework's applicability is related to the
    numerical approximation of the approach. We need to obtain long trajectories in simulation, but this
    may lead to poor quality in the gradients obtained for loco-manipulation problems \cite{antonova_rethinking_2022}.
    Moreover, the relaxations on which differentiable simulators rely on, may also introduce artifacts.
    Such complications need to be considered carefully and so in future work and
    the use of sampling or proxies for loss functions could be investigated as a possible mitigation.
    Finally, in this work, for interpretability's sake, only simple policies have been tested out.
    In future real applications,
    to increase the complexity of interaction, we aim to
    replace them with neural networks as differentiable function approximations, possibly generating richer motions.
    
    \printbibliography

\end{document}